\documentclass[journal]{IEEEtran}
\usepackage{graphicx}
\usepackage{amsmath}
\usepackage{times}
\usepackage{epsfig}
\usepackage{amssymb}
\usepackage{epstopdf}
\usepackage{float}
\usepackage{multirow}

\usepackage[font=small,labelfont=bf]{caption}

\begin{document}
\title{Order-aware Convolutional Pooling for Video Based Action Recognition}

\author{Peng Wang,
		Lingqiao Liu,
        Chunhua Shen,
        and Heng Tao Shen%
\thanks{P. Wang and H. T. Shen are with School of Information Technology and Electrical Engineering,  University of Queensland, Australia (email: p.wang6@uq.edu.au;
shenht@itee.uq.edu.au).}%
\thanks{L. Liu is with School of Computer Science, University of Adelaide, Australia (email: lingqiao.liu@adelaide.edu.au).}

\thanks{C. Shen is with School of Computer Science, University of Adelaide, Australia (email: chunhua.shen@adelaide.edu.au).}
\thanks{P. Wang's contribution was made when visiting University of Adelaide.}%
}

\markboth{Order-aware Convolutional Pooling for Video Based Action Recognition}%
{Wang \MakeLowercase{\textit{et al.}}: Order-aware Convolutional Pooling}
\maketitle

\begin{abstract}

Most video based action recognition approaches create the video-level representation by temporally pooling the features extracted at each frame. The pooling methods that they adopt, however, usually completely or partially neglect the dynamic information contained in the temporal domain, which may undermine the discriminative power of the resulting video representation since the video sequence order could unveil the evolution of a specific event or action. To overcome this drawback and explore the importance of incorporating the temporal order information, in this paper we propose a novel temporal pooling approach to aggregate the frame-level features. Inspired by the capacity of Convolutional Neural Networks (CNN) in making use of the internal structure of images for information abstraction, we propose to apply the temporal convolution operation to the frame-level representations to extract the dynamic information. However, directly implementing this idea on the original high-dimensional feature would inevitably result in parameter explosion.
	To tackle this problem, we view the temporal evolution of the feature value at each feature dimension as a 1D signal and learn a unique convolutional filter bank for each of these 1D signals.  We conduct experiments on two challenging video-based action recognition datasets, HMDB51 and UCF101; and  demonstrate that the proposed method is superior to the conventional pooling methods.

\end{abstract}


\IEEEpeerreviewmaketitle

\tableofcontents

\section{Introduction}

\begin{figure*}[t]
\begin{center}
\captionsetup{justification=centering}
\includegraphics[scale=.6]{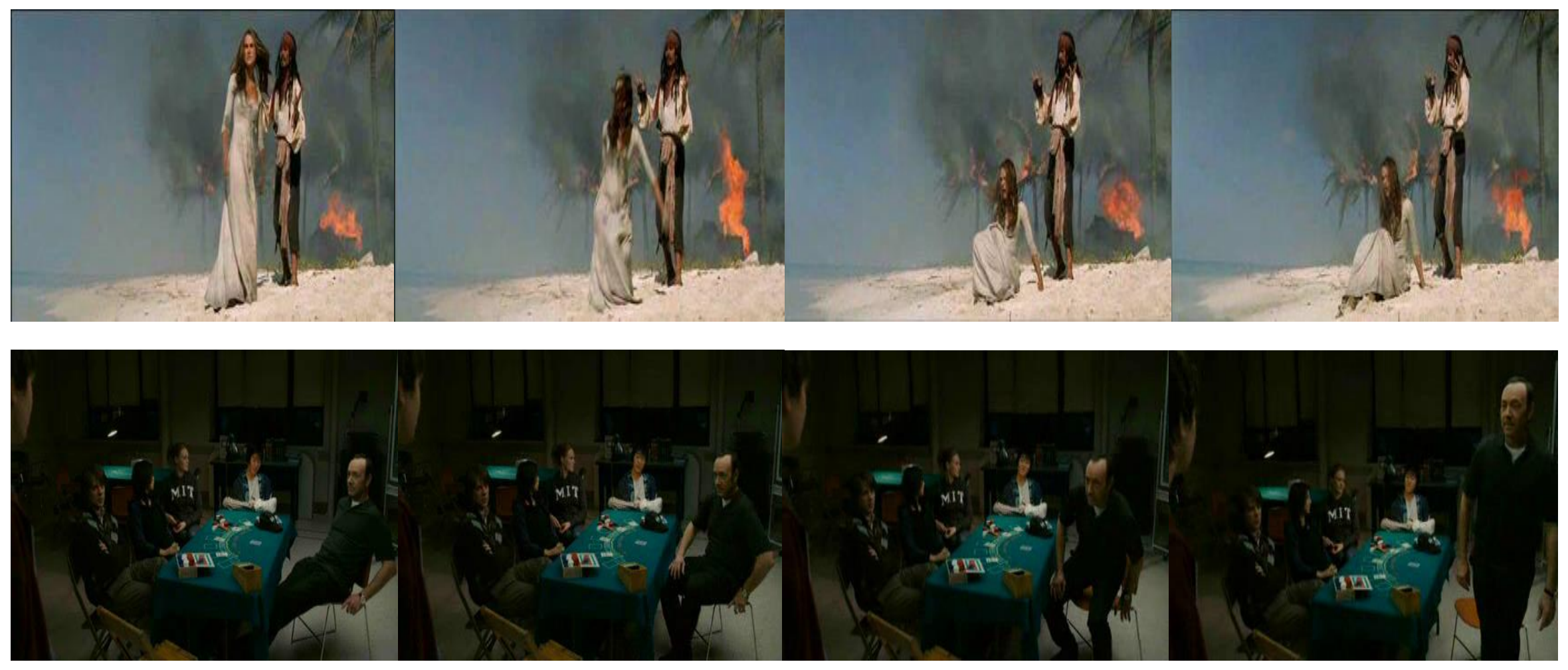}
\end{center}
   \caption{Order-preserving sample frames from two actions. Upper: Sit. Lower: Stand up.}
\label{fig:intro}
\end{figure*}

\IEEEPARstart{A} video is composed of a sequence of frames and the frame sequence reflects the evolution of the video content. Thus, a video can be naturally represented by a sequence of frame-level features which may describe either the visual patterns or motion patterns at a specific time step. To generate a vectorized video representation, a common practice is to apply temporal pooling, e.g., average or max pooling, to the frame-level features. However, these temporal pooling methods are problematic because they completely ignore the frame order and consequently lose the dynamic information contained in the temporal domain. In other words, the same pooling result will be obtained after randomly shuffling the frames. The frame-order, however, plays an important role in identifying actions or events in the video because it unveils the evolution of the video content. Fig.~\ref{fig:intro} shows some sampled order-preserving frames of two videos describing ``sit" and ``stand up" respectively. As can be seen, the frame order reflects the trend of the actions and encodes crucial discriminative information for distinguishing these two actions. A remedy to direct temporal pooling is to adopt temporal pyramid pooling which coarsely considers the temporal structure of a video by partitioning a video into a set of segments and deriving the representation of the video by concatenating these segment-level representations. It, however, still undergoes the loss of local dynamic information within each segment.

\begin{figure}[t]
\begin{center}
\captionsetup{justification=centering}
\includegraphics[scale=.35]{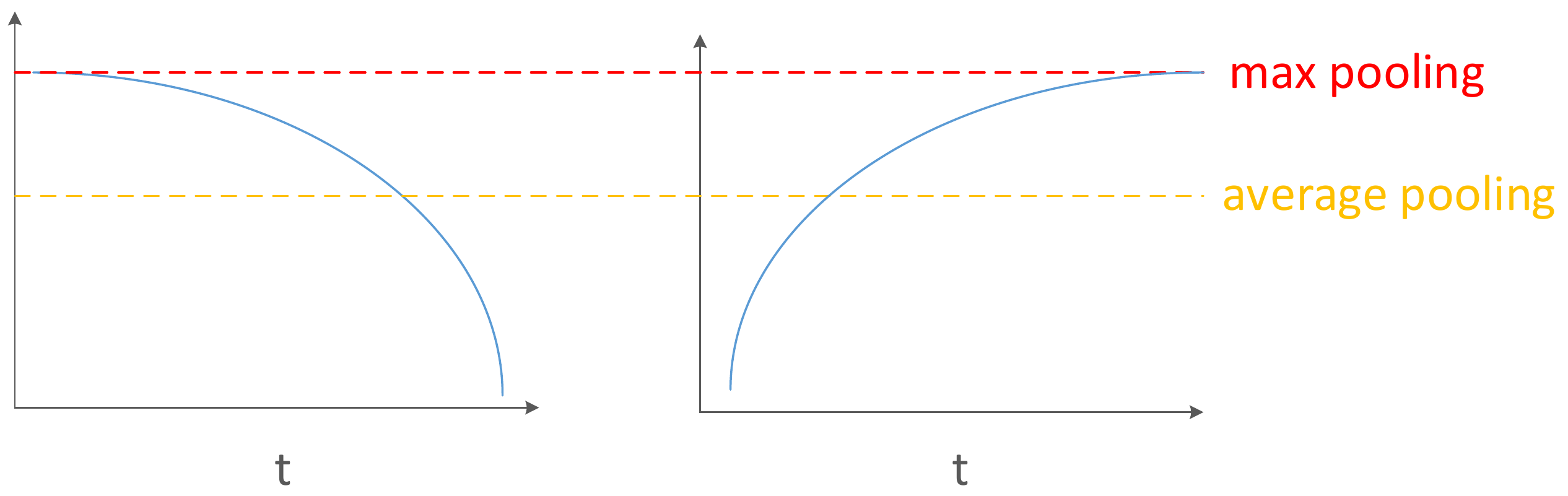}
\end{center}
   \caption{Illustration of average pooling and max pooling for 1D signal.}
\label{fig:intro0}
\end{figure}

To better capture the frame order information for action recognition, we propose a novel temporal pooling method to aggregate the frame-level representations. Our method is inspired by the use of the convolutional neural network (CNN) for image classification. In image classification, a CNN applies convolution operation to the local regions of the image to extract some discriminative visual patterns and uses pooling to make the representations invariant to some variations. Similarly,
a video can be regarded as an image with the image height being one, the image width being the length of the video and the number of image channels being the dimensionality of the frame-level features. Then we can apply convolution to the videos in the temporal domain to extract some discriminative patterns contained within the local temporal interval. However, when the frames are represented by high-dimensional features, such as CNN features or high-dimensional encodings (e.g., Fisher vector) of motion features, directly implementing this idea will lead to parameter explosion. The reason is twofold: (1) The number of parameters per filter equals to the frame-feature dimensionality times the size of local temporal interval; (2) for high dimensional frame features, a large number of filters will be needed to capture the useful information.
For example, when the frame-feature dimensionality is 10,000 and the interval size is 5 frames, 4,000 filters may be needed and this setting will result in about 200 million parameters in such a convolutional layer.
Training a network with such a large number of parameters will incur overwhelming computational cost and increase the risk of over-fitting especially when a limited number of training videos are available.

To address this issue, we propose to inspect the video from an alternative way, that is, we treat the temporal evolution of the feature value at each dimension of the frame-level features as a 1D signal. And the key of our method is to learn a set of filter banks for the 1D temporal signals in a supervised fashion. The filter bank for each feature dimension is unique and it servers as detectors to identify the discriminative local temporal patterns along the 1D temporal signal. After applying the filter banks to all the 1D signals, their filter responses are aggregated via a conventional temporal pooling method, i.e. average-pooling, max-pooling or temporal pyramid pooling to obtain the video-level representation.
Our method is advanced to the conventional pooling methods like max or average pooling since the latter ones only capture some simple statistics e.g., max value or direct current (DC) component of the 1D signals. As illustrated in Fig.~\ref{fig:intro0}, these two 1D signals cover opposite temporal information but conventional pooling methods will obtain the same pooling results because they have the same max value and DC value. In comparison, the proposed method can distinguish these two 1D signals by learning a filter to look into the local evoluation trend. Also, compared with the straightforward implementation which learns a filter with all frame-feature dimensions involved, the proposed method strategy significantly reduces the number of model parameters while still being able to capture the frame order information. For example, when the feature dimensionality and interval size are 10,000 and 5 respectively and 3 filters are adopted for each dimension, the number of parameters reduces to about 150,000 which is far less than that required in the  straightforward implementation. By conducting experiments on two challenging video-based action recognition datasets, HMDB51 and UCF101, we demonstrate that the proposed method achieves superior performance to the conventional pooling methods.

The rest of this paper is organized as follows: Section \ref{related work} reviews the previous work on action recognition and temporal feature pooling. Section \ref{our approach} elaborates the proposed order-aware pooling method. The experimental evaluation is presented in Section \ref{experiment}. Finally, Section \ref{conclusion} concludes this paper with discussions on future research.

\section{Related work} \label{related work}
\noindent \textbf{Video Based Action Recognition}. Inferring the action or event category from videos is a challenging problem that has attracted a lot of attentions. A considerable proportion of these works focus on designing some handcrafted features as video representations. Early works represent the videos by first detecting some spatio-temporal interest points and extract local features around these points \cite{4587756, 5206779, DBLP:conf/bmvc/KlaserMS08, Laptev03space-timeinterest, Scovanner:2007}. Most of these mechanisms are extensions from 2D image to 3D video. By tracking points over time, the trajectory based representation was proposed. They obtain the trajectories either using KLT tracker \cite{Lucas:1981} or SIFT matching \cite{5206721}. Inspired by the dense sampling in image classification \cite{1467486}, Wang \textit{et al.} \cite{wang:2011} proposed the dense trajectory (DT). It tracks the densely sampled points using dense optical flow and extracts the local descriptors in the 3D domain along the trajectories. To explicitly overcome camera motion, the improved dense trajectory (IDT) was proposed \cite{Wang2013}. It uses human parts detected by human detectors to estimate the motions between consecutive frames. Another strategy utilized to improve the classification performance is that they replace bag-of-words encoding with Fisher vector encoding \cite{Perronnin:2010}. Based on DT or IDT, people take some further steps to investigate some relevant problems such as how to effectively fuse different descriptors \cite{6909477} or how to encode these descriptors \cite{peng14}.

Most of the aforementioned methods derive the video representation in an unsupervised fashion. Resorting to the supervision information,
people propose several mid-level representations such as subvolumes \cite{peng:stack}, attributes \cite{NIPS2014_5565}, action parts \cite{DBLP:conf/mm/LiangLC13}, salient regions \cite{6751447} or actons \cite{6751554}. Some methods train a classifier for each discriminative part and fuse the classification scores to get a video-level representation. Some other methods treat the mid-level representations as local features and encode them using Fisher Vector encoding to derive the global representation.

Recently, along with the success of deep learning in image classification, there are some efforts dedicated to applying deep learning to video based action recognition. In \cite{6165309}, Ji \textit{et al.}
apply 3D convolution over 3D volumes of videos to capture spatio-temporal information. To learn a better spatio-temporal deep model, Karpathy \textit{et al.}
\cite{KarpathyCVPR14} collect a large video dataset that contains one million video clips for network training. In \cite{arXiv:1412.0767} the authors collect another large-scale video dataset and propose a generic spatio-temporal features for video analysis based on 3D convolution. Since these 3D convolution based models do not benefit from models pre-trained on large-scale image dataset for video classification, Mansimov \textit{et al.} investigate how to initialize the weights in 3D convolutional layers using the weights learned from 2D images to boost video classification performance. To explicitly take into consideration both the appearance information and motion information for action recognition, Simonyan and Andrew propose a so-called two-stream CovNet \cite{Andrew14}. While the spatial stream adopts an image-based network
structure to extract appearance features from sampled frames, the temporal stream takes as input stacked optical flow to capture motion
information. The decision scores of these two streams are fused together for video classification. Apart from CNN, RNN with LSTM cells \cite{Hochreiter:1997:LSM} is employed to learn the long-range temporal dynamics. In \cite{Donahue_2015_CVPR}, each video frame is fed into a CNN and they place a LSTM layer on top of the fully connected layers of the CNN to predict the video class at each time step. And these predictions are averaged for final classification.

\noindent \textbf{Feature Pooling Methods.} Images or videos are usually represented by a set of local descriptors and pooling is adopted to aggregate the statistics contained in these local descriptors. Two standard pooling methods are average pooling and max pooling. Average pooling captures the DC component of the feature values at a given dimension.
On the contrary, max pooling concerns only the maximum value and ignores the count statistics.
To overcome these limitations, a so called Generalized Max Pooling \cite{Murray:2014:GMP} was proposed which equalizes the similarity between each local representation to the pooled representation via re-weighting the local representations. However, it may risk magnifying the influence of the noisy statistics.
The work in \cite{Fernando2015a} proposes to capture the evolution of the video content via learning a function that is able to preserve the temporal ranking of the frames. The parameters of this function are used as the representation for the video. Employed to capture the long-range temporal information contained in the video, LSTM based method can be regarded as a temporal pooling method as well. Its advance is that it uses gating cells to adaptively control when to forget the signal. At the same time, however, this design introduces in a large number of parameters which makes LSTM based methods not suitable for tasks with small-scale training data. Catering for first-person video classification, the work in \cite{Ryoo_2015_CVPR} proposes to combine several pooling methods, i.e. sum pooling, max pooling, histogram of change and gradients' pooling, together to capture the dynamic information. Although obtaining better performance in first-person videos, this method cannot generalize to general video analysis most of which are third-person videos.

\section{Our proposed pooling method}
\label{our approach}
\begin{figure*}[ht]
\begin{center}
\includegraphics[scale=.6]{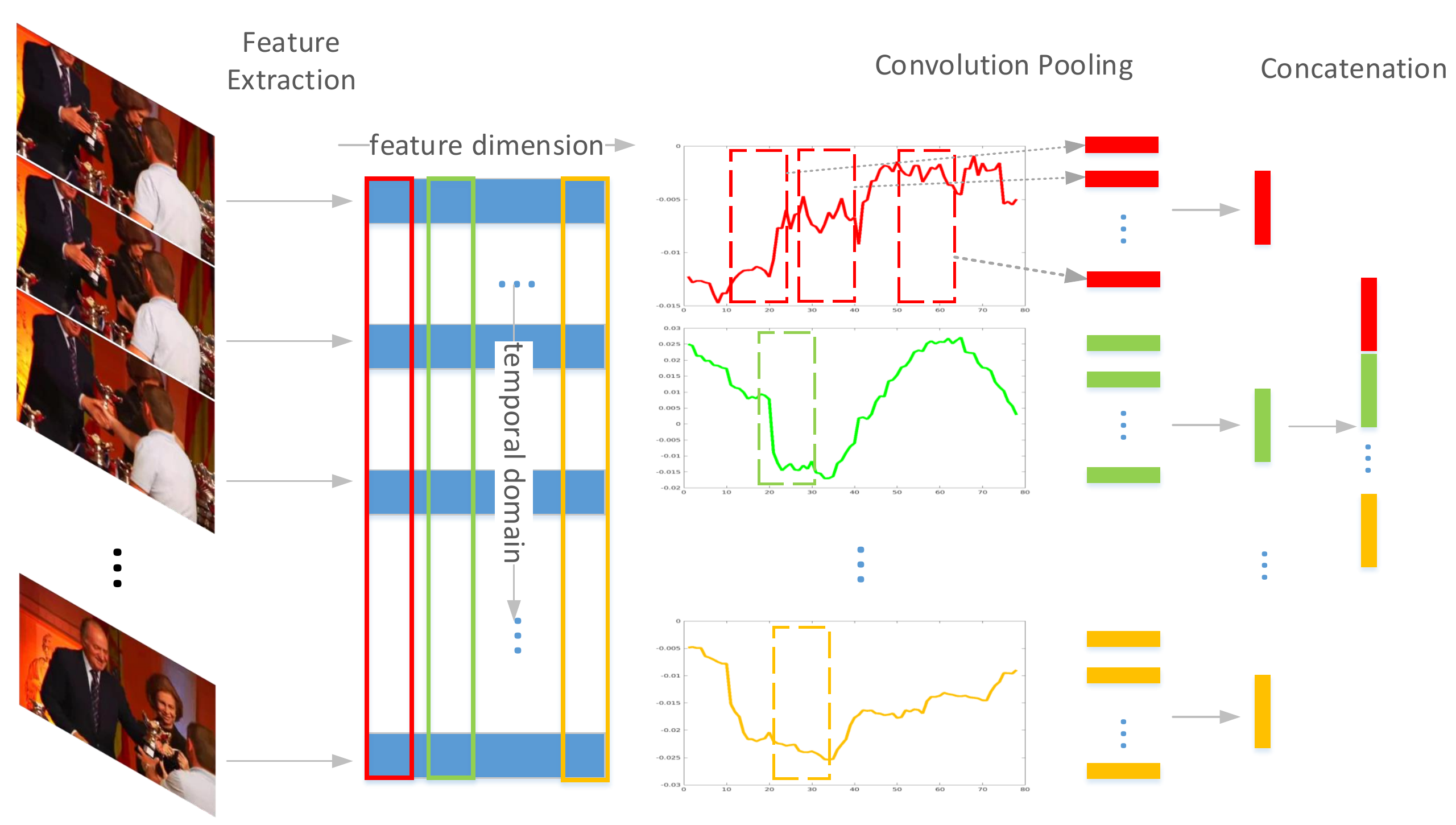}
\end{center}
   \caption{Illustration of order-aware pooling.}
\label{fig:network}
\end{figure*}

The general idea of the proposed order-aware convolutional pooling is shown in Fig.~ \ref{fig:network}. First, it extracts either appearance features or motion features from each video frame. Then a convolutional filter bank is learned and applied to each feature dimension over the temporal domain. The filter response signals of each dimension are aggregated as the dimension-level representation. Finally, these dimension-level representations are concatenated together as the video representation for classification.

\subsection{Frame-level representation preparation}
\label{Frame-level representations}

Appearance information and motion information embody different characteristics of the videos and they can compensate each other to describe a video. To take advantage of both information sources, we represent a video frame by concatenating both the appearance features and the motion features. We use CNN features of each video frame as the frame-level appearance features considering the proven success of CNN achieved in the field of image classification. Specifically, each frame is fed into a CNN model \cite{NIPS2012_4824} pre-trained on ImageNet \cite{imagenet_cvpr09} and the activations of the second fully-connected layer are used as its appearance features. For motion features, we resort to improved dense trajectory (IDT) \cite{Wang2013} considering its good performance for action recognition. Originally, IDT was proposed to generate a video representation which is obtained by aggregating the Fisher vectors of all the trajectories over the entire video by sum pooling. To create the IDT representation of a video frame, we first encode the trajectories passing this frame using Fisher vector encoding and then aggregate the coding vectors of these trajectories into a frame-level motion feature. The Fisher vector tends to be high dimensional which makes network training computational expensive. To address this problem, we adopt a supervised dimension reduction method \cite{DBLP:journals/corr/WangCSLS15} that is a variant of \cite{Liu:2013} to perform dimensionality reduction for the motion features. Compared with other methods like PCA, this method is very efficient in learning the dimensionality reduction function and performing dimensionality reduction especially in the scenario of reducing high-dimensional features to medium dimensions. Specifically, one only needs to calculate the mean of features in each class, which gives a data matrix $\mathbf{\bar S} \in \mathbb{R}^{c\times D}$, where $D$ indicates the feature dimensionality and $c$ indicates the total number of classes. Each column of $\mathbf{\bar S}$, denoted as $\mathbf{s}_i,~i= 1,\cdots,D$, is treated as a $c$-dimensional `signature' for the $i$-th feature. Then we perform $k$-means clustering on all $D$ `signatures' to group them into $k$ clusters, with $k$ being the target dimension. Thus the $D$ feature dimensions are partitioned into $k$ groups and this grouping pattern is used to perform dimensionality reduction.

\subsection{Order-aware convolutional pooling}

After feature extraction, a video is represented by a sequence of frame-level features. The objective next is to design a pooling method that can benefit from the frame order of the videos.
Recall that CNN makes use of the spatial structure of an image by applying convolution operation at different local regions of an image, our general idea is to apply convolution to the frame-level features over the temporal domain to make use of the 1D temporal structure (frame order) of the videos.

In image based CNN, a convolutional layer is composed of a filter bank and a nonlinear activation function. The filters in the filter bank can be regarded as some detectors which are applied to different local regions of an image to extract some discriminative patterns and non-linear function is applied to the responses of the filers to introduce nonlinearity to the neural network. Suppose the input of a convolutional layer are $K$ feature maps with the size of $H\times W$ from the previous convolutional layer, where $H$, $W$ denotes the height and width of the feature map respectively. A convolutional filter is operated on a small spatial support of the input feature maps, say, a small local region with the size of $h\times w$. For each local region, the convolutional layer computes $f(\left(\sum_{k=1}^K  \mathbf{W}_{k}^T \mathbf{r}^{k}_{ij}\right)+\mathbf{b})$, where $\mathbf{r}^k_{ij}$ denotes the flatten version of the activations within the $h\times w$ region at the $k$th feature map and its dimensionality will then be $h \times w$. The function $f(\cdot)$ is a non-linear function such as ReLU ($f(x)=max(0,x)$). The parameters $\{ \mathbf{W}_k \}$ and $ \mathbf{b}$ are to be learned during network training. Assuming the number of filters adopted in a convolutional layer is $n$, the total number of parameters of this layer is $w\times{h}\times{K}\times{n}+n$.

Inspired by the capability of CNN in making use of the spatial structure of the 2D image, we study applying convolution on action recognition via exploiting the temporal information of the videos. Recall that we represent a video as a sequence of frame-level feature vectors. For such a representation, we can treat our video representation as a special case of image representation with image height being 1, image width being the length of the video and the number of image channels being the dimensionality of the frame-level features. Then analogous to the convolutional operation adopted on images, we can learn some filters to extract the discriminative patterns within different temporal local regions. Suppose the dimensionality of the frame-level features is $K$, for the $t$th interval with the length being $l$, the convolutional operation computes $f(\left(\sum_{k=1}^K \mathbf{W}_{k}^T\mathbf{r}^{k}_{t}\right)+\mathbf{b})$ ($\mathbf{W}_{k}  \in \mathbb{R}^{l\times{n}} $), where $\mathbf{r}^k_{t}$ is a $l$-dimensional vector with its dimensions corresponding to the feature values at the $k$th dimension within the interval. Similar to the way of calculating the number of parameters in 2D images, the number of model parameters of such a convolutional layer will be $l\times{K}\times{n}+n$. Since the video frames are usually represented by high-dimensional features e.g., fully-connected CNN features or high-dimensional coding vectors, a large number of filters will be needed to capture the useful information and this will result in parameter explosion. Assuming that the interval size here is 8, the number of filters adopted is 4,000 and the dimensionality of the frame-level features is 10,000, the total number of parameters involved is about \textbf{320,000,000}. Training a model with such a huge number of parameters will incur prohibitively high computational cost as well as increase the risk of over-fitting.

To address this problem, in this work, we inspect the video from an alternative way. That is we treat the feature value evolution of one feature dimension over the temporal domain as a 1D temporal signal as shown in Fig.~\ref{fig:network} and represent a video as $K$ independent such 1D signals. The rationality behind is that for many high-dimensional features such as Fisher vector, the correlation between different feature dimensions tend to be small \cite{Zhang_2014_CVPR}. For each of such 1D signals, we learn a unique filter bank and similar to the 2D convolution at each convolution step these filter banks operate on a local temporal interval, that is, the filter output at time $t$ is calculated as $f(\mathbf{W}_{k}^T \mathbf{r}^{k}_{t}+\mathbf{b}_k)$. Similar to the 2D case, the term $\mathbf{r}^{k}_{t}$ denotes the vectorized representation of the $t$th interval at the $k$th feature dimension and its dimensionality equals $\l$, the size of the temporal interval.
In this case, since the filter bank is applied only to very low dimensional interval vectors, the number of filters required will be dramatically reduced,  e.g. reducing from 4000 to 3. Consequently, the number of model parameters will be far less than that involved in the aforementioned straightforward implementation. Let's assume that the number of filters for each dimension-wise filter bank is $\bar{n}$, then the total number of model parameters will be $l \times K \times \bar{n} +  K \times \bar{n}$. Assuming again that the interval size is 8, the number of filters adopted for each 1D signal is 3 and the dimensionality of the frame-level feature is 10,000, the total number of parameters involved will become about \textbf{240,000}, only being $1/1000$ of that in the straightforward implementation.

The output of the convolution operation of each 1D signal is a set of filter response vectors at different temporal locations. Obviously, the number of responses varies with the length of the videos. Considering that a fixed-length video representation is required for video classification, the pooling operation is employed to aggregate the varying number of response vectors of each feature dimension into a fixed-length dimension-level representation.

\begin{figure}[h]
\begin{center}
\captionsetup{justification=centering}
\includegraphics[scale=.5]{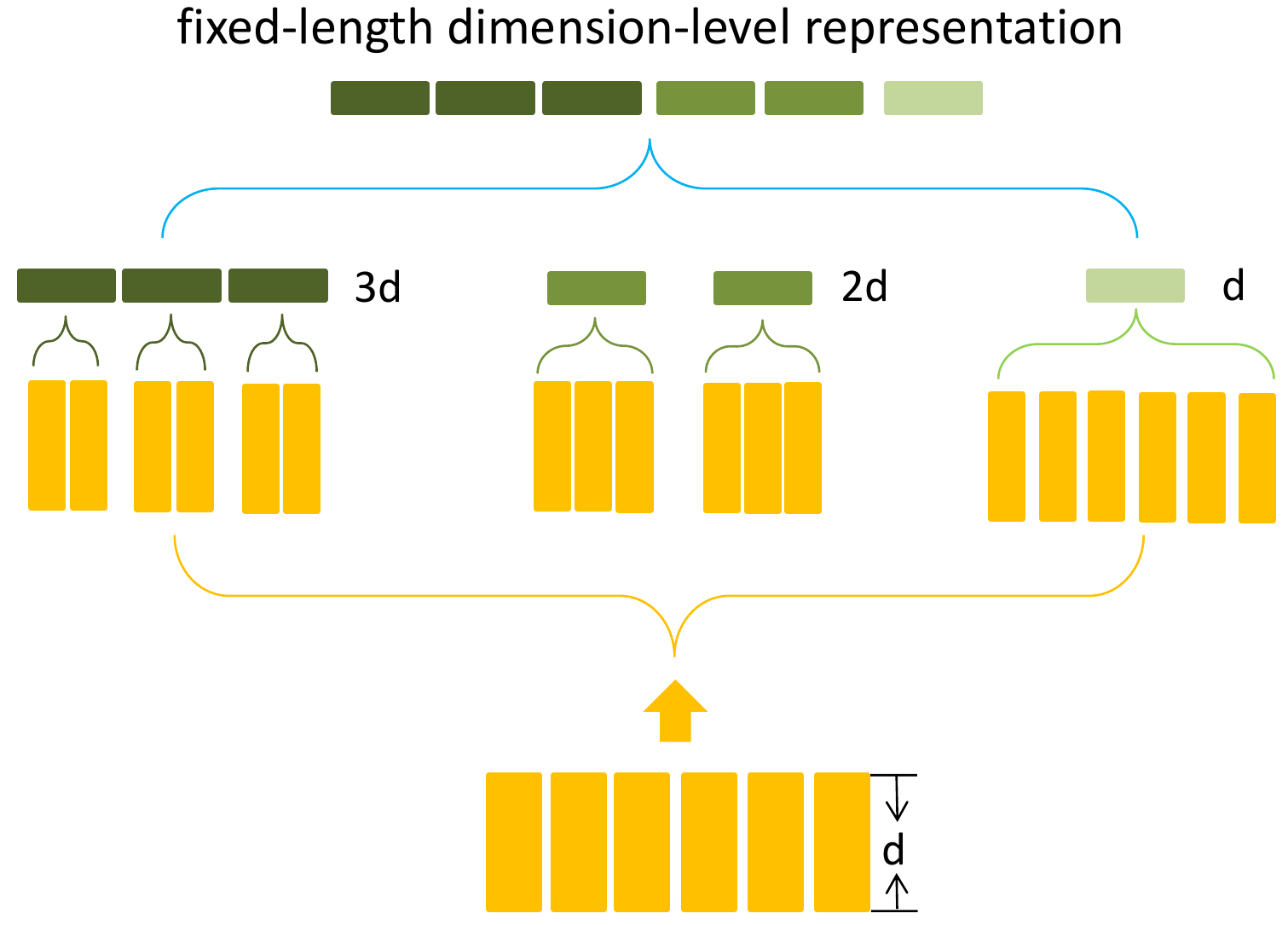}
\end{center}
   \caption{Illustration of temporal pyramid pooling.}
\label{fig:tp}
\end{figure}

To explicitly take into consideration the long-range temporal structure of the videos, we propose to use the temporal pyramid pooling to aggregate the local filter responses. Fig.~\ref{fig:tp} shows a three-level temporal pyramid pooling. The first level pools all the filter responses of a feature dimension directly over the temporal domain. For the $i$th level, the filter responses are partitioned into $m_i$ segments temporally and within each segment we perform max pooling. Then the representations of each segment will be concatenated together to from the representation for this dimension. So if the dimensionality of each segment-level representation is $d$, the dimensionality of the $i$th level will be $m_i\times d$ and the dimensionality of the dimension-level representation will be $d\sum_{i=1}^L m_i$, where $L$ is the number of levels used in the temporal pyramid pooling. After pooling the local responses, each dimension is represented by a fixed-length vector and the dimension-level representations are concatenated together to generate the representation of the video. Formally, the video representation can be expressed as follows:
\begin{equation}
\begin{aligned}
&\mathbf{P} = [\mathbf{P}^T_1, \mathbf{P}^T_2, \cdots, \mathbf{P}^T_k,\cdots, \mathbf{P}^T_K]^T,  \\
\text{where}, ~& \mathbf{P}^T_k = [\mathbf{P}^T_{k_{1}}, \mathbf{P}^T_{k_{2}}, \cdots, \mathbf{P}^T_{k_{L}}]^T,
\end{aligned}
\end{equation}
where $\mathbf{P}_{k_{j}}$ is the representation of the $j$th level of the $k$th dimension and $K$ is the dimensionality of the frame-level representation.

\subsection{Classification and learning of model parameters}
We learn the model parameters in a supervised fashion, that is, we add a classification layer on top of the outputs of the proposed pooling layer. It calculates $\mathbf{Y}=\varphi\left (\mathbf{W}_c\mathbf{P}+\mathbf{b}_c\right)$ where $\mathbf{W}_c$ and $\mathbf{b}_c$ are model parameters that will be learned during network training and $\varphi$ is the softmax \cite{MatCov} operation. The output $\mathbf{Y}$ is a probability distribution indicating the likelihood of a video belonging to each class. In the training stage, we use the following loss function to measure the compatibility between this distribution and ground-truth class label:

\begin{align}
	L(\mathbf{W},\mathbf{b})=-\sum_{i=1}^{N}\mathrm{log}(\mathbf{Y}(c_i)),
\end{align}
where $c_i$ denotes the class label of the $i$th video and $N$ is the total number of training videos. Recall that $\mathbf{Y}$ is a $\emph{c}$-dimensional vector and $c$ equals to the number of classes. Here we use $\mathbf{Y}(c_i)$ to denote the value at $c_i$th dimension of $\mathbf{Y}$. Using Stochastic Gradient Descent (SGD), in each step we update model parameters by calculating the gradient of an instance-level loss $L_i(\mathbf{W},\mathbf{B}) = -\mathrm{log}(\mathbf{Y}_{b}(c_i))$.

\section{Experimental evaluation}
\label{experiment}
\begin{figure*}
\begin{center}
\includegraphics[scale=.7]{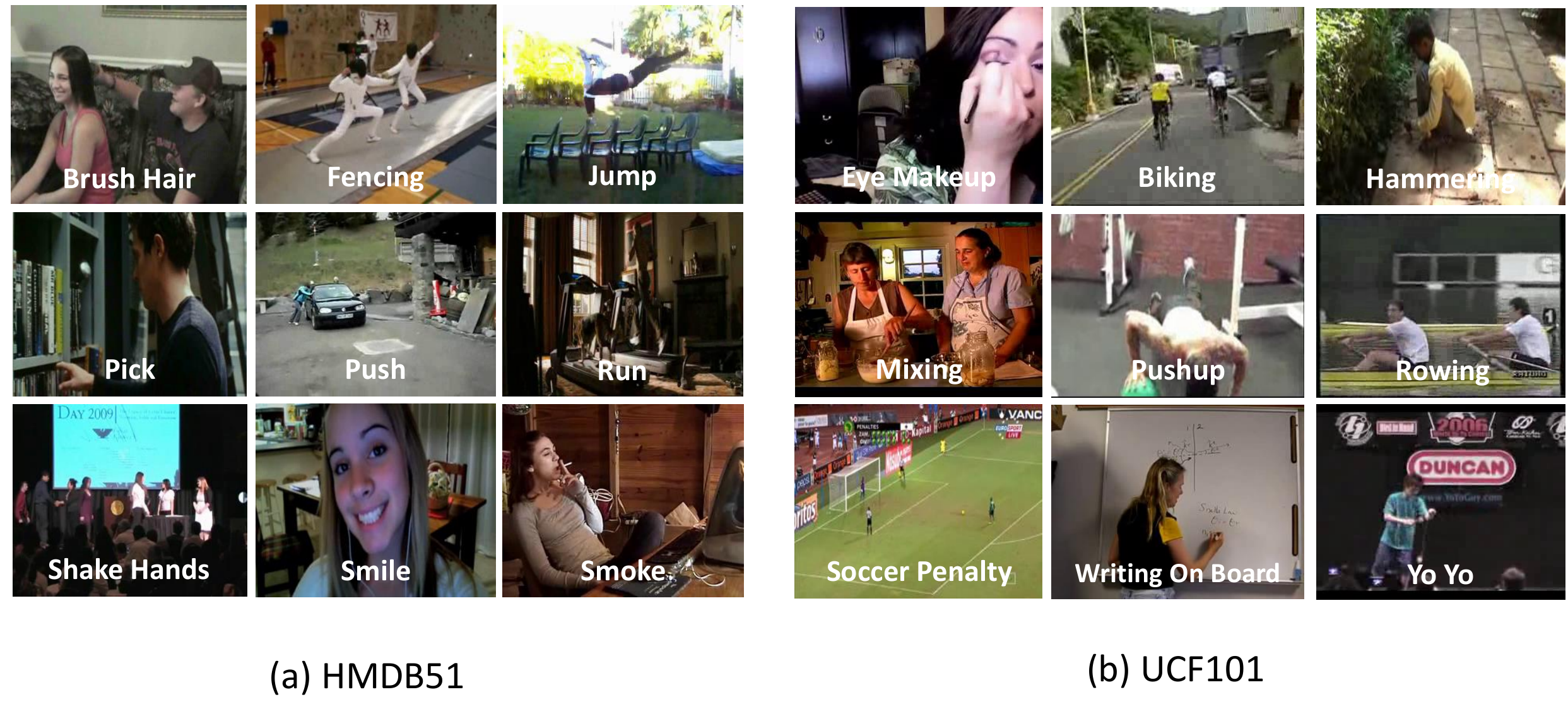}
\end{center}
   \caption{Example frames from (a) HMDB51 and (b) UCF101.}
\label{fig:sampled_frms}
\end{figure*}

The evaluation is performed on two datasets, HMDB51 \cite{Kuehne11} and UCF101 \cite{ucf101}. These two datasets are two of the most challenging datasets for video based action recognition. Fig.~ \ref{fig:sampled_frms}
shows some example frames of the two datasets.

\subsection{Experimental setup}
\subsubsection{Datasets}
The HMDB51 dataset \cite{Kuehne11} is collected from  various sources, such as movies, Prelinger archive and YouTube. It contains 6,766 video clips which are divided into 51 classes. According to the protocol in \cite{Kuehne11}, three training-testing splits are provided for this dataset. For each class, there are 70 training videos and 30 testing videos. The average classification accuracy over all the classes and splits is reported. This dataset provides two versions, a stabilized one and an unstabilized one. In our experiments, we use the latter version.

The UCF101 dataset \cite{ucf101} is composed of realistic action videos collected from YouTube. It contains 13,320 videos belonging to 101 classes. We use three train-test splits as in the standard evaluation protocol recommended by the dataset provider. For each class, the videos are split into 25 groups in which 7 groups are used for test and the rest are treated as training data. The classification
accuracy over all the classes and all the splits are reported as performance measurement.

\subsubsection{Parameter setting}
In the experiments, the parameters are set as follows unless otherwise stated. The interval size for the filters is set to be 8 and the interval stride is set to be 1. The number of filters adopted for each feature dimension is 3. The level of temporal pyramid is fixed to be 2 in the pooling layer when temporal pyramid pooling is employed. When SVM is applied in the experiments for classification, we fix $C=100$. Due to the redundancy between consecutive frames of a video, we sample 1 frame from every 5 frames for our method. Thus a filter in our method actually covers a range of 40 frames in the original video since the interval size of our filter is 8.

\subsubsection{Appearance feature} We utilize the 4,096-D activations of the second fully layer of AlexNet \cite{NIPS2012_4824} (a deep CNN model pre-trained on ImageNet) provided in Caffe \cite{jia2014caffe} as frame-level appearance features. Using better models such as ``vgg-deep" \cite{Simonyan15} as feature extractors can further boost the classification performance. However, for fair comparison with some existing methods \cite{Andrew14, KarpathyCVPR14} that are relevant to our work, we choose the shallower AlexNet model \cite{NIPS2012_4824}.

\subsubsection{Motion feature} We adopt the improved dense trajectory (IDT) \cite{Wang2013} as our frame-level motion features due to its good performance in action recognition. Different from \cite{Wang2013} that aggregates all the trajectories of a video into video-level representation using Fisher vector encoding, we aim at obtaining frame-level representation to make it compatible with the proposed network. To obtain the motion feature of a frame,
we consider the trajectories falling into a local neighbourhood of a frame with the size of the temporal neighbourhood being 10 (5 frames before and after a frame separately).
We encode these trajectories using Fisher vector coding with 256 Gaussians and the coding vectors are aggregated by sum pooling to form the frame-level representation. In this work, we extract HOF (108D) and MBH (196D) descriptors only to describe the trajectories. Since the Fisher vector is of high dimensionality (76800-D here) and this will make computation expensive or even infeasible. We adopt two treatments to address this problem. First, we only keep the ``mean" part of the Fisher vector and remove the ``covariance" part to reduce the dimensionality into half of its original implementation. Second, we use the dimensionality reduction method introduced in Section \ref{Frame-level representations} to reduce the dimensionality to 10,000.

\subsection{Performance evaluation}
In this subsection, the comparisons to the baselines on both appearance features and motion features will be given first to evaluate the effectiveness of the proposed pooling method. Then we investigate some other important aspects and/or properties related to our method, including the influence of the number of filters on classification performance and the complementarity between the proposed pooling method and the unsupervised global pooling used in IDT. Finally, we compare our method to the state-of-the-art.

\subsubsection{Comparison with baseline methods}
Both appearance feature and motion feature are employed to represent the video frames. In this subsection, we evaluate the efficacy of the proposed pooling method on these two types of features separately.

\noindent \textbf{Baselines for the appearance feature:} We now compare our method to the baselines using the frame-level appearance features. For the first two baselines, the frame-level CNN features are aggregated via average pooling (AP) or max pooling (MP) to obtain the video-level representations. After pooling, linear SVM \cite{REF08a} is employed for classification. For the third baseline, we adopt the pooling method proposed in \cite{Ryoo_2015_CVPR} which combines several pooling strategies together to capture the dynamic information. For the fourth baseline, temporal pyramid pooling (TP) is used to form the representation of a video, where max pooling is chosen to aggregate the frame-level features within each temporal segment and the configuration of TP is the same as that used in our method.

\noindent \textbf{Baselines for the motion feature:} Similar to appearance features, we apply average pooling, max pooling and temporal pyramid pooling to the frame-level motion features to create the video-level representations. The frame-level motion features are obtained in the same way as our method.

\begin{table}[t!]\normalsize
\caption{Comparison of the proposed pooling method to the baselines on HMDB51 using appearance information or motion information.}
  \centering
  \renewcommand{\arraystretch}{1.2}
    \begin{tabular}{l|p{2.6cm}c}
    \hline
    \multirow{4}{*}{Appearance} & AP & 37.5\% \\
    & MP & 36.5\% \\
    & PoT (no TP) \cite{Ryoo_2015_CVPR} & 36.5\% \\
    & TP & 39.2\% \\
    & Ours (MP) & \textbf{40.8\%} \\
    & Ours (TP) & \textbf{41.6\%}
    \\ \hline \hline
	\multirow{4}{*}{ Motion} & AP & 50.9\% \\
	& MP & 50.6\% \\
    & TP & 54.7\% \\
    & Ours (MP) & 52.8\% \\
    & Ours (TP) & \textbf{55.0\%} \\
    \hline \noalign{\smallskip}
    \end{tabular}%

  \label{baselinetab:HMDB51}%
\end{table}%

\begin{table}[t!]\normalsize
\caption{Comparison of the proposed pooling method to the baselines on UCF101 using appearance information or motion information.}
  \centering
  \renewcommand{\arraystretch}{1.2}
    \begin{tabular}{l|p{2.6cm}c}
    \hline
    \multirow{4}{*}{Appearance} & AP & 66.3\% \\
    & MP & 67.4\% \\
    & PoT (no TP) \cite{Ryoo_2015_CVPR} & 67.5\% \\
    & TP & 68.5\% \\
    & Ours (MP) & \textbf{69.3\%} \\
    & Ours (TP) & \textbf{70.4\%}
    \\ \hline \hline
	\multirow{4}{*}{ Motion} & AP & 80.0\% \\
	& MP & 80.2\% \\
    & TP & 81.6\% \\
    & Ours (MP) & 81.0\% \\
    & Ours (TP) & \textbf{82.1\%} \\
    \hline \noalign{\smallskip}
    \end{tabular}%

  \label{baselinetab:UCF101}%
\end{table}%

Tables \ref{baselinetab:HMDB51} and \ref{baselinetab:UCF101} demonstrate the results on HMDB51 and UCF101. From the results, we make the following observations:
\begin{itemize}
\item Motion features can lead to better classification performance comparing to appearance features. On both datasets, a method using motion features can outperform its counterpart that uses appearance feature by more than 10\%. This observation indicates that motion information plays a more important role in discriminating the actions in the videos.

\item On appearance features, the proposed pooling method can consistently outperform the baselines. In table \ref{baselinetab:HMDB51}, the network adopting max pooling outperforms AP, MP and TP by 3.3\%, 4.3\% and 1.6\% respectively. In Table \ref{baselinetab:UCF101}, our method with max pooling outperforms AP, MP and TP by 3\%, 1.9\% and 0.8\%. Note that the method in \cite{Ryoo_2015_CVPR} does not gain any improvement to max pooling which indicates that its pooling strategies e.g., histogram of change and gradients' pooling, are suited to first-person videos only. These results justify the advantage of the proposed network over direct pooling in capturing the dynamic information contained in the appearance features of the videos. Another observation is that after introducing temporal pyramid pooling into our network, the performance can be further boosted. Specifically, the classification accuracy is improved by 0.8\% on HMDB51 and 1\% on UCF101, which reveals the benefit of using temporal pyramid pooling in our method to capture the long-range information.

\item When working with motion features, our pooling method can obviously outperform AP and MP and achieve slightly better performance than TP. In Table \ref{baselinetab:HMDB51}, our method with max-pooling only gains improvement of 1.9\% and 2.2\% over AP and MP respectively. In Table \ref{baselinetab:UCF101}, our method with max-pooling outperforms AP and MP by 1\% and 0.8\% respectively. Again, these observations prove the importance of incorporating the frame order information. When temporal pyramid pooling is applied, performance of both the baseline methods and the proposed method are improved and our method obtains slightly better performance than TP on the two datasets. The advantage is not as significant as that on appearance features. This is probably due to
that the frame-level motion features have already encoded the dynamic information contained within adjacent frames, applying convolution on them cannot obtain significant improvement as on static appearance features.
\end{itemize}

\subsubsection{The impact of the number of filters}
The convolution operation constitutes the most important part of the proposed pooling method. In this part, we evaluate the impact of the number of filters. Specifically, we focus on investigating the change of classification performance w.r.t the number of filters adopted. We use frame-level appearance features and max pooling here. In our method, the interval size and interval stride are fixed to be 8 and 1 respectively and we choose three values 1, 3, 5 as the number of filters for each dimension.
Table \ref{tab:cov-neurons} shows the results.

As can be seen from the results on HMDB51, when only one filter is used the performance is unsatisfactory which means that one filter is insufficient to capture the dynamics contained in the given temporal interval of a feature dimension. Increasing $n$ can improve the performance and the best performance is obtained when $n=3$. After that, continuing to increase the number of filters leads to worse performance, which may be due to overfitting. On UCF101, using one filter produces worst performance again. However, unlike in HMDB51 the best performance is achieved when $n=5$. The reason may be that UCF101 has much more training data which makes the model training less prone to overfitting.

\begin{table}[h]\normalsize
 \caption{The impact of the number of filters $n$ in the proposed pooling method (4096-D frame-level appearance features are used for this evaluation.)}
  \centering
  \renewcommand{\arraystretch}{1.2}
    \begin{tabular}{l|cc}
    \hline
    \multirow{3}{*}{HMDB51} & $n=1$ & 38.9\% \\
    & $n=3$ & \textbf{40.8\%} \\
    & $n=5$ & 39.5\% \\
    \hline \hline
	\multirow{3}{*}{ UCF101} & $n=1$ & 67.8\% \\
    & $n=3$ & 69.3\% \\
    & $n=5$ & \textbf{69.6\%} \\
    \hline \noalign{\smallskip}

    \end{tabular}%

  \label{tab:cov-neurons}%
\end{table}%

\subsubsection{Performance on combined features}
In this part, we evaluate the performance of the proposed method when using both the aforementioned appearance features and motion features. More sepcifically, for each video frame we perform $L2$ normalization to the appearance feature and the motion feature respectively and concatenate the normalized features together as the frame-level representation. Since our method uses both the CNN features and IDT based motion features, it is fair to compare another baseline which concatenates the max-pooled frame-level CNN features and the global motion features. Here the global motion feature is obtained via aggregating the Fisher vectors of the trajectories over the entire video as in IDT\cite{Wang2013}.
Similarly, the CNN features and global motion features are $L2$ normalized separately before concatenation. Linear SVM is used for classification. Table \ref{tab:combined} shows the results. As can be seen from this table, our method can outperform the baseline by 1.6\% and 0.9\% on HMDB51 and UCF101 respectively. When combined with the unsupervisedly pooled motion features, our performance can be significantly boosted further. This observation shows that the representations learned by our method is strongly complementary to the representation obtained via an unsupervised pooling method.

\begin{table}[htbp]\normalsize
 \caption{Performance on combined features. For IDT, we use HOF and MBH only.}
        \centering

        \begin{tabular}{lcc}

            \hline\noalign{\smallskip}

                Methods  &    HMDB51  & UCF101 \\

            \noalign{\smallskip}

            \hline

            \noalign{\smallskip}

             CNN (max) + Global Motion Pooling   & 59.4\%          &   86.9\%                \\

             Ours             &  61.5\%         &	  87.8\%				\\
			 Ours + Global Motion Pooling  & \textbf{64.1\%} & \textbf{89.6\%} \\

            \noalign{\smallskip}

            \hline \noalign{\smallskip}

      \end{tabular}

      \label{tab:combined}

\end{table}

\subsubsection{Comparison with state-of-the-art}
In this part, we compare our method to the state-of-the-art methods on HMDB51 and UCF101. Note that, out performance can be further boosted by some strategies like employing a better CNN model \cite{Simonyan15} to extract CNN features or using higher dimensional motion features.

\noindent \textbf{HMDB51}
Table \ref{tab:overall-hmdb51} compares our method to several state-of-the-art methods on HMDB51. As can be seen, our method achieves the second best performance. Hybrid improved dense trajectories in \cite{peng14}, employs multiple unsupervised encoding methods i.e. Fisher vector \cite{Perronnin:2010}, VLAD \cite{VLAD} and LLC \cite{llc}. In comparison, our method is much more elegant in the sense that it relies on a single encoding module. Note that the best performed method, stacked Fisher vector, \cite{peng:stack} employs two-level Fisher vector encoding and concatenates them together as video representation. The work \cite{Andrew14} is a CNN based method and adopts frame sampling to handle the issue of video-length variation. The video evolution method \cite{Fernando2015a} captures the evolution of the content within the video via learning a ranking functions to rank frames.

\begin{table}[htbp]\normalsize
\caption{Experimental results on HMDB51.}
  \centering

  \renewcommand{\arraystretch}{1.1}
    \begin{tabular*}{8cm}{lc}
    \hline \noalign{\smallskip}
    Spatial-temporal HMAX network \cite{Kuehne11} & 22.8\% \\
    DT \cite{wang:2011} & 47.2\% \\
    Jain \textit{et al.} \cite{6619174} & 52.1\% \\
    DT+MVSV \cite{6909477} &55.9\% \\
    IDT \cite{Wang2013} & 57.2\% \\
    Hybrid IDT \cite{peng14} & 61.1\% \\
    Stacked Fisher Vector \cite{peng:stack} & \textbf{66.8\%} \\
    Two-stream ConvNet (average fusion) \cite{Andrew14} & 58.0\% \\
    Two-stream ConvNet (SVM fusion) \cite{Andrew14} & 59.4\% \\
    Video Evolution \cite{Fernando2015a} & 63.7\% \\
    Factorized Networks \cite{Sun_2015_ICCV} & 59.1\% \\
    Actionness \cite{Luo_2015_ICCV} & 60.4\% \\
    Ours & \textbf{64.1\%}
    \\ \hline \noalign{\smallskip}
    \end{tabular*}%

  \label{tab:overall-hmdb51}%
\end{table}%

\noindent \textbf{UCF101}
Table \ref{tab:overall-ucf101} shows the results on UCF101. We first compare our method to the LRCN \cite{Donahue_2015_CVPR} which utilizes the LSTM to aggregate the frame-level CNN features for action recognition. Our method outperforms it by 2\%. As can be seen in the lower part of Table \ref{tab:overall-ucf101}, our method performs best among these methods. The spatio-temporal convolution based method \cite{KarpathyCVPR14} performs worse than dense trajectory based methods \cite{6909477,Wang2013,peng14}. Our method can outperform two-stream CovNet \cite{Andrew14} by 1.6\%.
The Deep net \cite{Ng2015} stacks Long Short-Term Memory (LSTM) cells on top of a CNN for video classification. Still, we can achieve better performance than that.
\begin{table}[htbp]\normalsize
\caption{Experimental results on UCF101.}
  \centering

  \renewcommand{\arraystretch}{1.1}
    \begin{tabular*}{8cm}{lc}
    \hline \noalign{\smallskip}
    LRCN \cite{Donahue_2015_CVPR} (LSTM + CNN) & 68.2\% \\
    Ours CNN & \textbf{70.4\%} \\
    \hline \noalign{\smallskip}
    Spatio-temporal CNN \cite{KarpathyCVPR14} & 63.3\% \\
    DT+VLAD \cite{6909477} & 79.9\% \\
    DT+MVSV \cite{6909477} & 83.5\% \\
    IDT \cite{Wang2013} & 85.9\% \\
    Hybrid IDT \cite{peng14} & 87.9\% \\
    Two-stream ConvNet (average fusion) \cite{Andrew14} & 86.9\% \\
    Two-stream ConvNet (SVM fusion) \cite{Andrew14} & 88.0\% \\
    Deep Net \cite{Ng2015} & 88.6\% \\
    Factorized Networks \cite{Sun_2015_ICCV} & 88.1\% \\
    Ours & \textbf{89.6\%}
    \\ \hline \noalign{\smallskip}
    \end{tabular*}%

  \label{tab:overall-ucf101}%
\end{table}%

\section{Conclusions}
\label{conclusion}
We have proposed a novel temporal pooling method called order-aware convolutional pooling. It can well capture the dynamic information contained in the frame order while maintaining a tractable amount of model parameters. Experiments on two video-based action recognition datasets demonstrated the efficacy of our method. Note that apart from video classification, our method can be applied to other time-series tasks such as text classification where each word is represented by a feature vector.

\ifCLASSOPTIONcaptionsoff
  \newpage
\fi

\bibliographystyle{IEEEtran}
\bibliography{egbib}

\end{document}